# A Feature Clustering Approach Based on Histogram of Oriented Optical Flow and Superpixels

A.M.R.R. Bandara, L. Ranathunga
Department of Information Technology
Faculty of Information Technology, University of Moratuwa
Moratuwa, Sri Lanka.
ravimalb@uom.lk

N.A. Abdullah
Department of Computer System and Technology
University of Malaya
Kuala Lumpur, Malaysia.
noraniza@um.edu.my

# A Feature Clustering Approach Based on Histogram of Oriented Optical Flow and Superpixels

A.M.R.R. Bandara, L. Ranathunga
Department of Information Technology
Faculty of Information Technology, University of Moratuwa
Moratuwa, Sri Lanka.
ravimalb@uom.lk

N.A. Abdullah
Department of Computer System and Technology
University of Malaya
Kuala Lumpur, Malaysia.
noraniza@um.edu.my

*Abstract*—Visual feature clustering is one of the cost-effective approaches to segment objects in videos. However, the assumptions made for developing the existing algorithms prevent them from being used in situations like segmenting an unknown number of static and moving objects under heavy camera movements. This paper addresses the problem by introducing a clustering approach based on superpixels and short-term Histogram of Oriented Optical Flow (HOOF). Salient Dither Pattern Feature (SDPF) is used as the visual feature to track the flow and Simple Linear Iterative Clustering (SLIC) is used for obtaining the superpixels. This new clustering approach is based on merging superpixels by comparing short term local HOOF and a color cue to form high-level semantic segments. The new approach was compared with one of the latest feature clustering approaches based on K-Means in eight-dimensional space and the results revealed that the new approach is better by means of consistency, completeness, and spatial accuracy. Further, the new approach completely solved the problem of not knowing the number of objects in a scene.

I. INTRODUCTION

Segmentation of semantic objects is a crucial first step in video understanding applications such as annotating and indexing of videos for video search and retrieval purposes. High-level object recognition typically builds upon this step. As such, finding robust solutions for segmenting of objects in videos has become an active research problem. Standard approaches to the problem include motion structure analysis and appearance-based methods. Several recent studies have shown a significant improvement on using a flow of visual features for object segmentation [1]–[4].

Earlier studies such as in [5] showed that motion and 3D world structure can be used to segmentation of objects in videos. The semantic segmentations solely based on motion-derived 3D structures tend to neglect the objects with smaller regions. The study in [5] has shown that appearance-based methods can be used to detect some of the objects which are neglected by the motion and structure cues. Recent studies [1], [2] have showed that long-term flow analysis can help to accurately refine object regions. However, practically some of the objects may appear or disappear within a small fraction of the video due to the camera motion hence long term analysis may not feasible in some cases. The fast segmentation approach in [3] works reliably when satisfying certain conditions such as the object should move differently from its surrounding background in a good fraction of the video and at least a part of the object should move somewhere in the video. Such assumptions prevent the use of the algorithm for the majority of the videos due to uncertain behaviors of the salient objects. A previous study [4] has showed that a simple clustering technique can be used to cluster salient dither pattern (SDPF) features [6] by using temporal information and dither patterns, for semantic segmentation. However, it requires a human interaction for estimating the number of clusters (K) by inspecting a short sample of the video. Further, the algorithm tends to neglect some objects due to the inconsistent estimation error of K along the video. Further, the K should be re-estimated in cases of applying the algorithm to totally new set of videos.

This paper presents an improved unsupervised visual feature clustering approach for segmenting static and moving objects in videos with camera motion, by using short-term flow information and an appearance-based cue. Particularly the approach refines a set of estimated segments to form high-level object segments by using color based cue and local motion information of each of the estimated segments. It has been shown that superpixel can be used for estimating video segments as an initial step of segmenting high-level objects[3], [7]. The comparative study mentioned in [8] has revealed that the Simple Linear Iterative Clustering (SLIC) superpixel has the state-of-the-art performance by means of time efficiency, memory efficiency and segmentation reliability. In this new approach, SLIC superpixel is used to estimate low-level segments in a weighted Lab color space. SDPF is used as the point feature due to its invariant properties [6] and the optical flow is calculated using the pyramidal implementation of Lucas-Kanade optical flow [9]. The segment refining is done by using a short term histogram of oriented optical flow (HOOF) with Radial Basis Function (RBF) kernel on Bhattacharya distance [10] and mean colors of superpixels. A refining process is employed as a bottom-up approach by considering only the neighborhood relationships, starting from the superpixel extraction to the clustering of features correspond to the objects. The superiority of the new approach is demonstrated by comparing with the results in [4].

This work is financially supported by National Research Council, Sri Lanka under the grant number 12-017.

The rest of this paper is organized as follows. In section II, related studies are briefly discussed with respect to the feature clustering. Section III describes the new approach to feature clustering. Section IV describes the experimental setup for comparing the new approach with the previous works. Section V, the experimental results are discussed with followed by conclusion Section VI.

## II. RELATED STUDIES

In this section SLIC superpixels, SDPF feature and HOOF are described with respect to feature clustering.

### A. SLIC Superpixel

Superpixels capture redundancy in an image and greatly reduce the complexity of subsequent image processing tasks. They have proved increasingly useful for semantic segmentation [3]. SLIC superpixels are obtained by performing a local clustering of pixels in the 5-D space defined by the L, a, b values of the CIELAB color space and x,y pixel coordinates. In order to self-contain the paper, SLIC algorithm will be explained briefly here onward.

At the beginning of the algorithm, it places a given number of approximated superpixels at nearly equal grid intervals but with avoiding pixels that have higher gradient such that the pixels which are not located on edges. This step ensures the approximation of superpixels with minimum noise. Then a localized k-means clustering is employed by using the summation of the Euclidean distance of colors in CIELAB color space and spatial distance which is weighted by a spatial proximity emphasizing factor [8]. SLIC shows a strong perceptual homogeneity within the superpixels and O(n) time complexity hence it is used in this new approach.

### B. SDPF Points

This study mainly focused on clustering SDPF feature points with respect to high-level semantic object segments. SDPF is a point like low dimensional feature which is based on the concept in Compacted Dither Pattern Code (CDPC), which the dithering reduces both color and spatial data density while preserving the visual impression [11]. The adaption of the concept in CDPC by including both color and spatial information ultimately creates a point-like feature that extracted based on its saliency over a set of local dither patterns. The detailed steps of extracting SDPF features and its strong invariant properties are described in[4], [6]. The ultimate goal of the feature clustering is to recognize multiple objects by using the low dimensional SDPF descriptor hence SDPF points is used to initiate the optical flow estimation. Besides the main goal of this work, the geometrical invariant properties of SDPF is used to overcome the heavy geometrical transformation of object segments caused by the ego-motion.

### C. HOOF

The concept of HOOF feature was introduced in [10] for human activity recognition. The results of the same study have shown a time series of HOOF can characterize a high-level object motion. HOOF can be built by binning each flow vector according to its primary flow angle and weighted according to its magnitude. Practically the flow estimation of feature points is very noisy especially when the magnitude of flow is very low. Theoretically HOOF feature is robust to the flow noises due to the flow angles are weighted by the flow magnitudes. As described in [10], HOOF does not reside in Euclidean space hence similarity matching should be done by using a proper matching function from the space of histogram to high-dimensional Euclidean space.

Although this study is not related to activity recognition, the ability of HOOF to characterize specific motion is taken as an advantage in this approach. This approach argues that neighbors of low-level segments that belong to the same semantic object have similar HOOF features over a short period of time.

## III. PROPOSED CLUSTERING APPROACH

As the first step of the clustering, SDPF points are extracted from the first frame. Then calculate the flow using the pyramidal implementation of Lucas-Kanade optical flow estimation approach with taking the SDPF points as the references for the tracking. The optical flow is estimated for a predefined number of frames ($T_f$) while recording the flow angle and magnitude of each tracking point for each frame. The $T_f$ was set to 3 in the experiment, in order to assess the performance of the clustering approach with a shallower history of motion.

Once the flow estimation process reaches to the $T_f$, the clustering process is initiated by using SLIC superpixel algorithm. The original SLIC algorithm requires the input image in CIELAB color space due to two colors in CIELAB space with smaller Euclidean distance are perceptually similar. The $L$ component in CIELAB color space closely matches the human perception of brightness [12]. Therefore, the component L is very sensitive to shadows and different amount of light intensities over the video frame. The impact of this nature of L component causes the SLIC algorithm to adhere some pixels as an edge although the pixels are actually placed on boundaries of shadows but not actual object boundaries. Therefore the weighted Lab color such that $L'ab$ is obtained using (1).

$$L'ab = \begin{bmatrix} m & 0 & 0 \\ 0 & 1 & 0 \\ 0 & 0 & 1 \end{bmatrix} \begin{bmatrix} L \\ a \\ b \end{bmatrix} \quad (1)$$

The value of $m$ is kept at 0.5 in the experiment because the increment of the value of α caused the SLIC to detect many false boundaries.

SLIC algorithm further requires two other parameters, the number of clusters ($K$) and minimum distance for color proximity ($N_c$). In all the experiments conducted in this study, $K$ and $N_c$ were set to 50 and 10 respectively, based on the frame size of the videos in the dataset. A color cue is used in addition to the motion cue in a later step of the process of refining the superpixels. This color cue is obtained by averaging the colors of pixels in each superpixel segments. In the refining process, this average color similarity between two superpixel regions is used as one of the criteria for the refining. The similarity of the colors is obtained by using the Euclidean distance as defined in (2).

$$D_c=\sqrt{(L'_1-L'_2)^2+(a_1-a_2)^2+(b_1-b_2)^2} \quad (2)$$

Where $L'$, $a$ and $b$ are the weighted Lab components of average color calculated for a superpixel.

The next step is to obtain HOOF for the flow of SDPF points in each superpixel. Therefore, a time series of HOOF features is calculated for each superpixel using the feature points located within the region of the superpixel. A flow vector, $v=[x,y]^T$ with the angle $\theta=\tan^{-1}\frac{y}{x}$ is in the range $-\pi<\theta\leq\pi$. By mapping the range of the $\theta$ to $0<\theta\leq2\pi$ the corresponding bin $b$ can be calculated by (3).

$$b=\left\lfloor\frac{\theta}{B}\right\rfloor \quad (3)$$

where the $B$ is the number of bins. In the experiments, the value of $B$ is set to 30 as it did on its original work [10]. The contribution of the vector to the HOOF is calculated by (4).

$$h_b=h_b+\sqrt{x^2+y^2} \quad (4)$$

where the $h_b$ is the value of $b^{th}$ bin in the histogram $h$. The HOOF for each superpixel is calculated by considering the local flow vectors correspond to the each of the superpixels. The time series of the HOOF can be calculated by using the recorded flow data within the feature points. At the end of the calculation each of the superpixel will have a set of HOOF features which's the size is equal to the predefined length of time series.

As described in [10] HOOF cannot be treated as a simple vector in Euclidean space hence Euclidean-based distance measurements cannot be applied to compare two HOOF features. Hence, the similarity of histograms is measured by calculating a kernel based on Riemannian metric on a hypersphere which is created by the square root representation of the histograms. The kernel is defined as (5).

$$k(h_1,h_2)=\sum_{i=1}^{B}\sqrt{h_{1,i}h_{2,i}} \quad (5)$$

where $h_1$ and $h_2$ are the two histograms. In this study, the above kernel was implemented as mentioned in [10] using RBF kernel on Bhattachcharya distance between histograms, as shown in (6).

$$k_{RBF}(h_1,h_2)=\exp\left(-d(h_1,h_s)\right) \quad (6)$$

where $d(h_1,h_s)$ is Bhattachcharya distance between $h_1$ and $h_2$ histograms. Note that $h_1$ and $h_2$ are normalized to form probability density functions. The range of RBF kernel $k_{RBF}$ is [0, 1] where 1 means exactly same and 0 means less similar.

Since it is needed to compare the similarity between two time series of HOOF, first, the RBF kernel is calculated for each pair of histograms created in the same instance of time from neighboring superpixels. Then the magnitude of the set of RBF kernel values is calculated to measure the similarity between whole time series of HOOF by (7).

$$D_h=\sqrt{\left(k_{RBF}^{t=1}\right)^2+...+\left(k_{RBF}^{t=T_f}\right)^2} \quad (7)$$

Where $D_h$ is the similarity measure between the two time series of HOOF and $k_{RBF}^{t=1}$ is the RBF kernel of two HOOFs calculated for two superpixels at the time step 1 and so on.

*A. Refining Process of Superpixels*

The regions of high-level semantic objects are formed by refining the superpixels. The refining process is involved in following 6 steps.

1. Consider the first superpixel and calculate $D_h$ and $D_c$ with all its 4-connected neighbors.
2. Check the merging criteria ($D_h > T_h$ or $D_c < T_c$) where $T_h$ and $T_c$ are predefined thresholds.
3. In case the merging criteria are satisfied, check whether, either of the current superpixel or the neighbor superpixel has been given a cluster number.
4. If either of superpixels has a cluster number, then the cluster number is shared by both superpixels and merge the time series of HOOF and average colors of the two superpixels.
5. If neither of superpixels has a cluster number, then create a new cluster number by incrementing the number of existing clusters, assign to both superpixels and merge the time series of HOOF and average colors of the two superpixels.
6. Then go to the next superpixel, calculate $D_h$ and $D_c$ with all its 8 neighbors and repeat from the step 2, until there is no superpixel remains without a cluster number.

The major problem of merging superpixels is to select merging criteria which accurately quantify the similarity between two superpixels, which represents the likelihood of being the two superpixels in a single semantic object. Due to the high color complexity of typical objects, the merging criteria should be not limited to appearance based cue. Therefore in this approach, the merging criteria have been designed to quantify the similarity by at least one of the color or motion cues at any given instance.

In order to find the two thresholds $T_h$ and $T_c$, initially the maximum possible value of $T_h$ such that 3 was assigned to $T_h$ and checked the correctness of cluster merging manually, by varying $T_c$. The observations suggested that $T_c$ can be varied effectively within a very narrow range such that [0, 30] due to the weighting process of CIELAB colors. In order to balance the color proximity measurement, $T_c$ was set to 15 and check the merging correctness manually, by varying $T_h$. The results suggested that 1 and 15 are good enough for $T_h$ and $T_c$ respectively for most of the time. Hence, the same values were used for all the experiments conducted in this study.

## IV. EXPERIMENTAL SETUP

The performance of the new approach is evaluated using the CamVid dataset [13]. The dataset contains high-resolution video with a frame rate of 30Hz, captured from the perspective of a driving automobile. The video contains both the moving and static objects. The performance evaluation requires measuring the correspondence between the clusters to the semantic objects in a scene hence spatial centroid of each of the clusters were used to find the correspondence. If a centroid of a cluster is laid within the boundary of an object, then it is considered that the particular cluster belongs to the same object. The assessment was done for 100 visually different frames comparing with the ground truth video.

*A. Evaluation Criteria*

One of the objectives of this study to assess the improvement of the new clustering approach compared to the previous clustering approach described in [4]. Therefore, the same evaluation criteria are used as in the previous work. In order to self-contain the paper, the performance measurement used in both of the studies are briefly described here onward. An objective comparison for each frame is done using a spatial accuracy measure $S_{er}$ [14], as defined in (8).

$$S_{er} = \frac{\sum_{n=1}^{N}|M^{cl}(n) \oplus M^{ref}(n)|}{\sum_{n=1}^{N}|M^{ref}(n)|} \quad (8)$$

where $M^{cl}(n)$ is the set of points clustered into the $n^{th}$ object, $M^{ref}(n)$ is the set of points that have manually clustered to the $n^{th}$ object, i.e. the ground truths, $N$ is the number of semantic objects in the frame and $\oplus$ denotes the exclusive disjunction of the two sets.

The completeness of the clustering $S_{compl}$ is defined in (9) where the symbol $\cap$ denotes conjunction of the two sets.

$$S_{compl} = \frac{\sum_{n=1}^{N}|M^{cl}(n) \cap M^{ref}(n)|}{\sum_{n=1}^{N}|M^{ref}(n)|} \quad (9)$$

## V. EXPERIMENTAL SETUP

The performance evaluation results of the new clustering approach are presented comparing with the results in [4]. In the previous study, the completeness measurement and spatial accuracy of three clustering approach in 8-dimensional space (flow angle, flow magnitude, four colors in the salient dither patterns and x,y coordinates) using K-means (K-Means-8D), Expectation Maximization (EM-8D) and density based scan (DBScan) were evaluated. In this paper, the performance analysis of the worst approach in the previous study, which used the DBScan is omitted in order to simplify the visualization of the results of the new approach. Here onward the new approach which is based on Refining Superpixels using HOOF and Color cue will be referred as RSHC. Fig. 1 shows the obtained results for RSHC with comparing the K-Means-8D and EM-8D.

Fig. 1 demonstrates that the new approach has obtained a small but significant improvement over the previous methods. Precisely RSHC achieved over 0.87 of average completeness for the whole video where as K-Means-8D and EM-8D achieved around 0.86 and 0.81 of average completeness respectively. Although the improvement in average completeness is not very significant, Fig. 1 evident that RSHC achieved a significant improvement by means of consistency over the frames. The consistency of the K-Means-8D and EM-8D are lower because the less robustness of the estimation

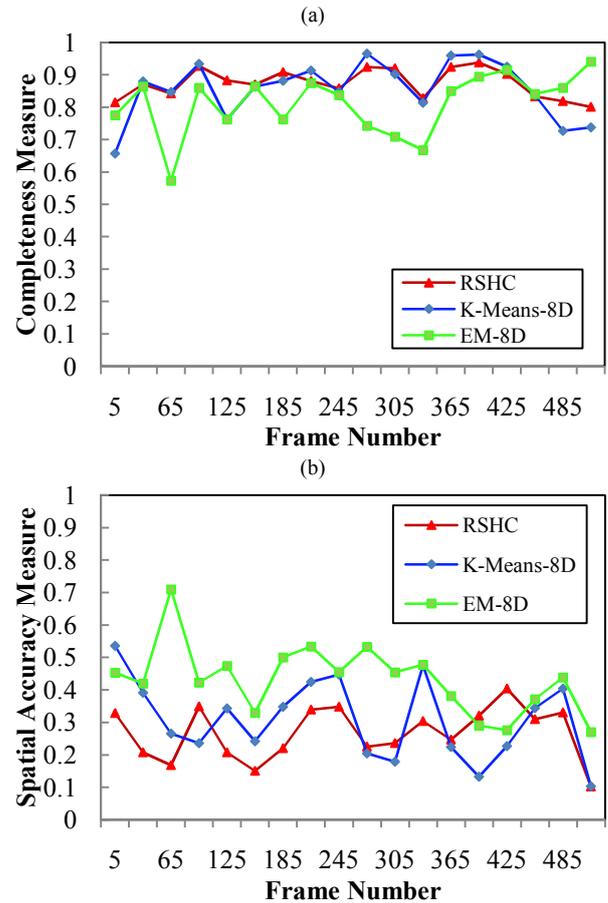

Fig. 1. Comparison of the three approaches with the sequence 16E5. (a) Completeness measure. (b) Spatial accuracy measure.

method of a number of clusters. As RSHC does not require a prior knowledge of a number of objects in the scene, indeed it showed a good consistency over different scenes. A sample

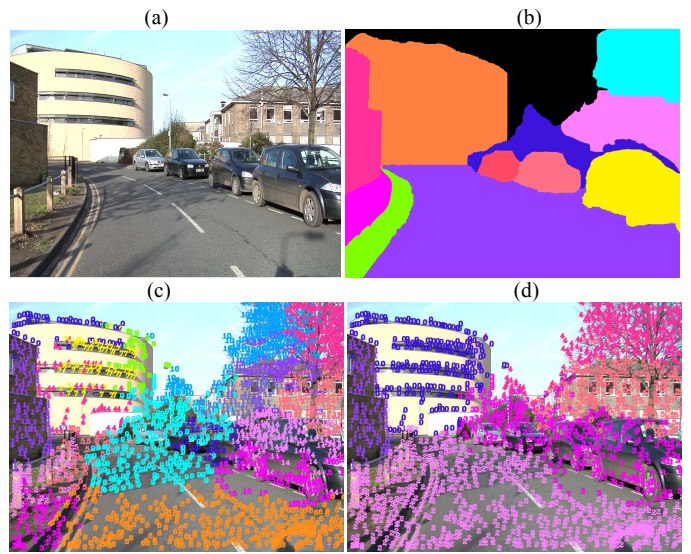

Fig. 2. A sample of clustering SDPF points using K-Means-8D and RSHC (Frame No. 4). (a) Original. (b) Ground-truth segments (c) Resultant frame after clustering with K-Means-8D (d) Resultant frame after clustering with RSHC

instance of clustering with K-Means-8D using the estimated number of clusters based on the number of feature points, and a sample instance of clustering with RSHC are shown in Fig. 2 in order to demonstrate the quality of the performance of the new approach. The Fig. 2 clearly shows that RSHC is less affected by the lack of information i.e. number of objects in the scene whereas it tends to heavy over-clustering in the K-Means based approach.

Besides the completeness and consistency, RSHC also shows a significant improvement on spatial accuracy error compared to the other two methods. The average spatial accuracy error of RSHC method is around 0.27 where the K-Means-8D and EM-8D are 0.31 and 0.42 respectively. We believe that the improvement in spatial accuracy error is gained by the ability of accurate adherence to object boundaries by SLIC algorithm with the weighted CIELAB colors.

## VI. CONCLUSION

This paper presents a new visual feature clustering approach on clustering SDPF feature points to semantic objects by refining SLIC superpixels using both short-term time series of HOOF and color similarity measurements. The study has proved that the proposed scheme works consistently over different visual scenes in a video and can perform better than the methods which apply global clustering with indirect estimation of the amount of clusters. The performance evident that time series of HOOF can be used to characterize both static and moving objects. However, since HOOF does not include the spatial distribution of flow angle and magnitude within a superpixel, the issue how to incorporate motion boundaries for further improving the spatial accuracy of clustering, remains. The findings of this study will be used for detecting and recognizing multiple semantic objects using SDPF feature descriptor.